\documentclass[a4paper,conference]{IEEEtran}

\usepackage[T1]{fontenc}
\usepackage{lmodern}
\usepackage{cite}
\usepackage{amsmath,amssymb,amsfonts}
\usepackage{algorithmic}
\usepackage{graphicx}
\usepackage{xcolor}
\usepackage{subcaption}
\usepackage{booktabs}
\usepackage{tabularx}
\usepackage{url}
\usepackage[hidelinks]{hyperref}
\usepackage[capitalize]{cleveref}
\usepackage{orcidlink}

\crefname{section}{Sec.}{Secs.}
\Crefname{section}{Section}{Sections}
\crefname{figure}{Fig.}{Figs.}
\Crefname{figure}{Figure}{Figures}
\crefname{table}{Tab.}{Tabs.}
\Crefname{table}{Table}{Tables}

\begin{document}

\title{\LARGE \bf
Performance Evaluation of an Integrated System for Visible Light Communication and Positioning Using an Event Camera
}

\author{%
\IEEEauthorblockN{
Ryota Soga\IEEEauthorrefmark{1}, 
Masataka Kobayashi\IEEEauthorrefmark{1}, 
Tsukasa Shimizu\IEEEauthorrefmark{2}, 
Shintaro Shiba\IEEEauthorrefmark{3},\\
Quan Kong\IEEEauthorrefmark{3}, 
Shan Lu\IEEEauthorrefmark{1}, 
Takaya Yamazato\IEEEauthorrefmark{1}
}
\IEEEauthorblockA{\IEEEauthorrefmark{1}Graduate School of Engineering, Nagoya University, Nagoya, Japan\\
\texttt{\{soga,kobayash,lu,yamazato\}@yamazato.nuee.nagoya-u.ac.jp}}
\IEEEauthorblockA{\IEEEauthorrefmark{2}TOYOTA MOTOR CORPORATION, Toyota, Japan\\
\texttt{tsukasa\_shimizu@mail.toyota.co.jp}}
\IEEEauthorblockA{\IEEEauthorrefmark{3}Woven by Toyota, Inc., Tokyo, Japan\\
\texttt{\{shintaro.shiba, quan.kong\}@woven.toyota}}
}

\maketitle

\begin{abstract}
% ...
\end{abstract}

\begin{abstract}
Event cameras, with their high temporal resolution and high dynamic range, can function as visual sensors comparable to conventional image sensors. These characteristics enable event cameras to excel in detecting fast-moving objects and operating in environments with extreme lighting contrasts, such as tunnel exits. As a result, they are gaining attention as a next-generation sensing technology for autonomous vehicles.

In this study, we propose a novel self-localization system by integrating visible light communication (VLC) and visible light positioning (VLP) functionalities into a single event camera. The system uses VLC to obtain coordinate information from transmitters, while VLP estimates the distance to each transmitter. By combining these two types of information, the receiver can determine its own location, even in environments where GPS signals are unavailable, such as tunnels.

The most significant contribution of this study lies in the simultaneous realization of both VLC and VLP functions using a single event camera. Multiple LEDs are installed on the transmitter side, each assigned a unique pilot sequence based on Walsh-Hadamard codes. The event camera uses these codes to calculate the presence probability of each LED within its field of view, allowing for clear separation and identification. This enables high-capacity data transmission via MISO (Multi-Input Single-Output) communication and accurate distance estimation through triangulation using POC (Phase Only Correlation) between multiple LED pairs. To the best of our knowledge, this is the first vehicle-mounted system to achieve simultaneous VLC and VLP using an event camera.

We conducted performance evaluations in real-world conditions by mounting the system on a vehicle traveling at 30 km/h (8.3 m/s). The results showed that the root mean square error (RMSE) of distance estimation was within 0.75 m for ranges up to 100 meters, and the bit error rate (BER) for communication remained below 0.01 across the same range.
\end{abstract}

\begin{figure}[ht]
    \centering
    \includegraphics[width = 0.9\linewidth]{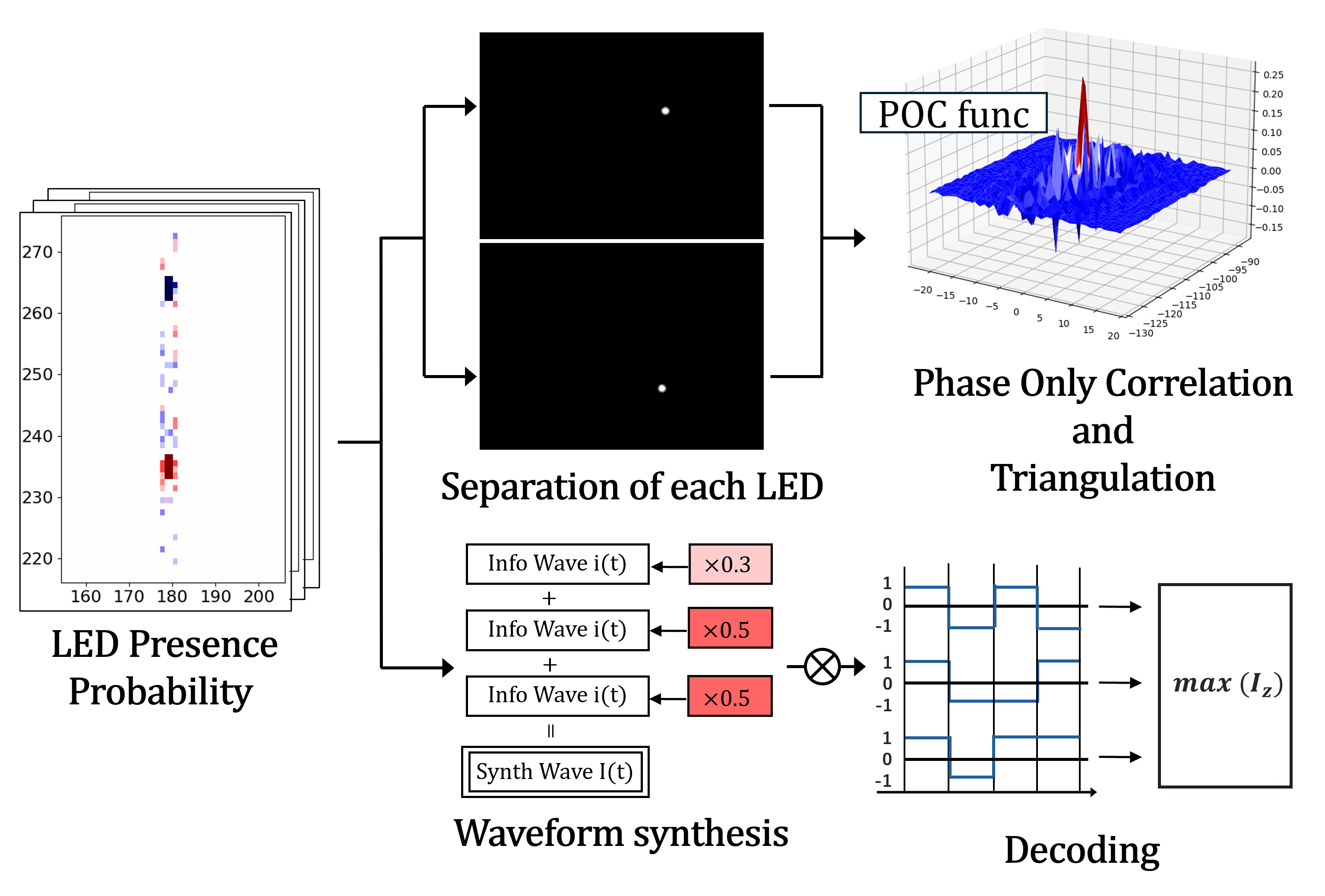}
    \vspace{-3mm}
    \caption{A general overview of the proposed model}
    \vspace{-7mm}
    \label{fig:eye-catcher}
\end{figure}
\vspace{-2mm}
\section{Introduction}
\label{sec:introduction}
\vspace{-1mm}
Smart-city initiatives are accelerating worldwide to digitalize urban infrastructure and optimize mobility and logistics.\cite{smartcity} Streetlights, traffic signals, and signage are being repurposed as sensing/communication nodes, enabling data exchange among vehicles, pedestrians, and drones to form a mobility data backbone. City-scale cooperative control, however, requires each vehicle to localize itself autonomously and reliably and align with the infrastructure’s coordinate frame.

Currently, integrated systems combining LiDAR, GNSS (Global Navigation Satellite System), and INS (Inertial Navigation System) are primarily being investigated~\cite{lio-sam,multilidar}. In particular, high-precision localization is achieved by matching the approximate location information obtained via GNSS with 3D shape data captured by LiDAR and pre-constructed 3D maps.

However, GNSS relies on direct waves from satellites and environments where line-of-sight is difficult to maintain, such as tunnels, underground spaces, or areas with densely packed buildings, there is a significant degradation in localization accuracy ~\cite{gpsafit}~\cite{gpstunnel}. These limitations suggest that GNSS-based systems may not be sufficient for future autonomous applications, such as underground delivery infrastructure being explored in countries like Switzerland
~\cite{switzerland}. To address this issue, this study proposes a novel localization system that does not rely on radio waves, utilizing an event camera combined with visible light communication (VLC) and visible light
positioning (VLP).
All visible lights from streetlights, traffic signals, signage, vehicles, and drones can be utilized for data transmission and localization.

An event camera is a sensor that asynchronously outputs events (coordinates, time, and polarity of luminance change) based on the luminance changes of individual pixels. By accumulating events within a given time window, the camera can also provide frame-based images similar to those from conventional image sensors. The event camera's primary advantage lies in its sparse output, which enables low-data transmission while achieving extremely high temporal resolution~\cite{Gallego2022pami}. These characteristics allow for embedding more information in the time domain for VLC and realizing high scan rates for precise distance measurements in VLP.

In our previous work, we independently evaluated vehicle-based VLC systems~\cite{soga2025} and VLP systems~\cite{kobayashi}. However, for practical deployment, it is necessary to implement both functions simultaneously with a single transmitter–receiver setup. While some prior studies have reported integrated VLC-VLP systems using high-speed cameras~\cite{ohmura}, to our knowledge, none have implemented such systems using event cameras. Moreover, existing works that explore VLC~\cite{ACM2019} or VLP~\cite{vlp1,vlp2} using event cameras have mainly evaluated performance in static or low-speed pedestrian environments, and there is a lack of studies addressing vehicle-mounted scenarios at moderate speeds, specifically up to 30 km/h. In this context, we present what we believe to be the first integrated VLC-VLP system using an event camera for vehicular applications. Moreover, our goal is to achieve, using event cameras, communication rates and ranging accuracy equal to or exceeding those of the previously mentioned high-speed-camera-based integrated VLC-VLP system.

To realize this system, several technical challenges must be addressed. The first challenge lies in the difference between the blinking patterns traditionally used in VLC and VLP. A simple time-division multiplexing approach would reduce the scanning frequency for VLP and the communication rate for VLC. Therefore, a unified blinking pattern that can be shared by both functions is required. The second challenge concerns improving the accuracy of VLP. 
Previous studies have reported significant variation in distance measurement results, likely due to electronic noise inherent to event cameras. Overcoming this requires a robust technique that enhances the precision of distance estimation.

To address these challenges, we propose extending the pilot sequence used for LED separation in VLC to VLP as well. Moreover, by using three or more LEDs, we achieve both the unification of blinking patterns and enhanced measurement accuracy. Specifically, we estimate the presence probability distribution of LEDs by computing the cross-correlation between the known Walsh-Hadamard (WH) code-based pilot sequences and the data received at each pixel. In VLC, this distribution allows for LED identification in MISO communication. In VLP, POC (Phase Only Correlation)~\cite{poc} is applied to the presence probability distribution to estimate the distance between LEDs. By using three or more LEDs and averaging the distances obtained from multiple pairs, we enhance the stability of the measurement results.

To evaluate the effectiveness of the proposed system, we conducted outdoor vehicle-mounted experiments. In terms of communication performance, we achieved a bit error rate (BER) of less than 0.01 at distances of 30 - 100 m, with vehicle speeds up to 30 km/h (8.3 m/s). This performance level indicates that error-free communication is achievable with the application of error correction codes. In terms of distance measurement, we achieved a root mean square error (RMSE) of less than 20 cm at approximately 50 - 60 m, with a scan rate reaching approximately 300 Hz.

The main contributions of this research are as follows:
\begin{itemize}
    \item We proposed a method to integrate VLC and VLP functions by sharing a common pilot sequence and applying POC, thereby constructing a vehicle-mounted VLC-VLP integrated system using an event camera. This approach enables simultaneous operation of both functions while maintaining a high scan rate of approximately 300 Hz.
    \item By using three or more LEDs, we introduced multi-point distance measurement and averaging techniques that reduced variation in the VLP results, achieving high-precision distance estimation with RMSE less than 20 cm at 50 - 60 m.
\end{itemize}
\vspace{-3mm}

\section{System Model}
\label{sec:system_model}
\vspace{-1mm}
\begin{figure*}[tbp]
\centering
\includegraphics[width=0.75\linewidth]{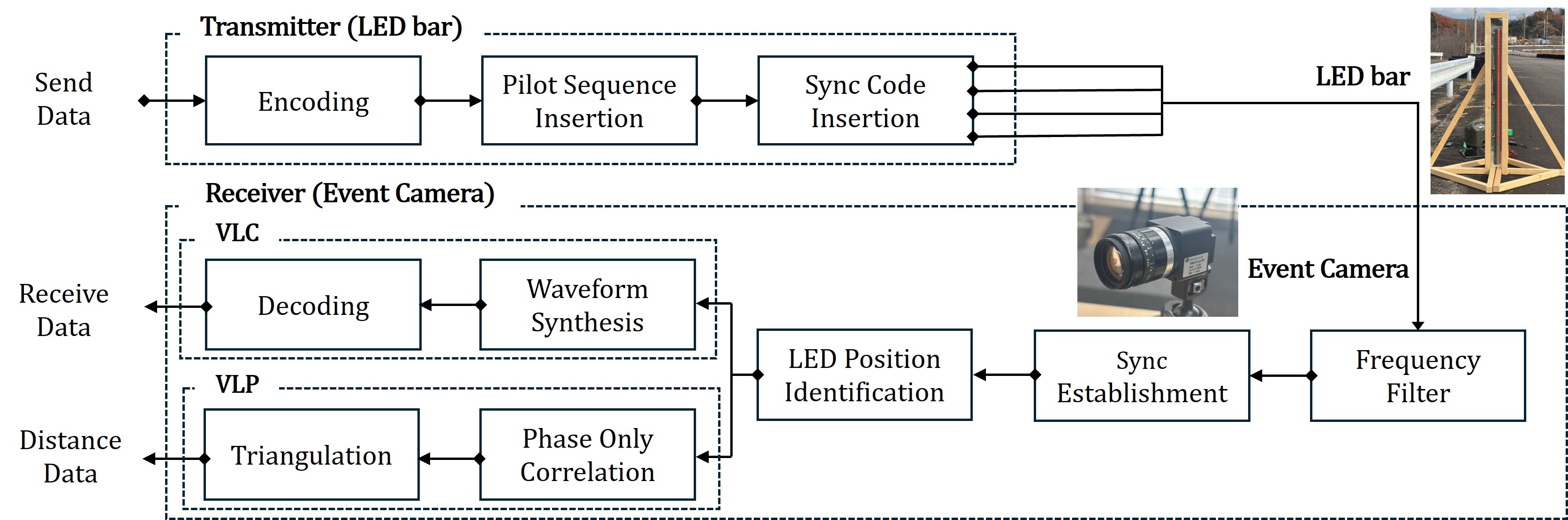}
\caption{System model}
\vspace{-6mm}
\label{fig:systemmodel}
\end{figure*}

The system model proposed in this study is illustrated in Fig.~\ref{fig:systemmodel}. On the transmitter side, a vertically aligned LED bar consisting of 96 red LEDs driven at a blinking frequency of 10 kHz is used. These LEDs are grouped into multiple adjacent units, referred to as LED Clusters. Each Cluster transmits the same waveform, allowing the signal to be easily recognized even from long distances. Meanwhile, different Clusters transmit different waveforms. The adopted modulation scheme is On-Off Keying (OOK), which is simple and allows for fast processing.

On the receiver side, we use the "SilkyEvCam" equipped with the SONY IMX636 event-based image sensor. This camera has a resolution of 1280×720 and offers high temporal resolution of less than 100 µs. Since event cameras output asynchronous events triggered by luminance changes at each pixel, they are particularly well-suited for detecting high-frequency blinking light sources such as the transmitter used in this system. We use Metavision Studio for data acquisition and enable the hardware-level "event frequency high-pass filter," which allows only high-frequency events to pass through. This effectively suppresses low-frequency events from background illumination, highlighting only the signal components from the transmitter. Additionally, a Region of Interest (ROI) is configured to restrict the number of active pixels within the field of view, thus preventing the event throughput of the SilkyEvCam from being exceeded and ensuring stable operation.

The received data is processed through multiple steps. First, a Frequency Filter is applied to extract high-frequency components and suppress background noise, enabling rough localization of the LED bar.

Next, temporal synchronization between the transmitter and receiver is established. A Barker code is used as the synchronization signal, and by sliding it along the time axis of the received data and computing the autocorrelation, the time at which the correlation peak appears is identified as the synchronization point \cite{soga2025}.

After synchronization, a common processing step is performed for both VLC and VLP. The field of view is divided into spatial regions, and in each region, the received signal is cross-correlated with a known pilot sequence to estimate the presence probability distribution of each LED Cluster. This distribution is then used in VLC to separate the signals from each Cluster and decode the corresponding transmitted waveforms. In the case of VLP, the presence probability is used to isolate events originating from each Cluster. These are then converted into 2D arrays, to which POC is applied to estimate the inter-cluster pixel distance. Based on the obtained pixel distances, triangulation is performed to accurately estimate the distance between the transmitter and receiver~\cite{kobayashi}.

Through the above processing flow, the proposed VLC-VLP integrated system leverages the unique characteristics of event cameras to enable both high-precision communication and distance estimation between the transmitter and receiver. In the following sections, we provide a detailed explanation of the core algorithms developed in this study.
\vspace{-2mm}
\subsection{LED presence probability distribution}
\label{method:led-probability}
\begin{figure}[tbp]
\centering
\includegraphics[width=0.80\linewidth]{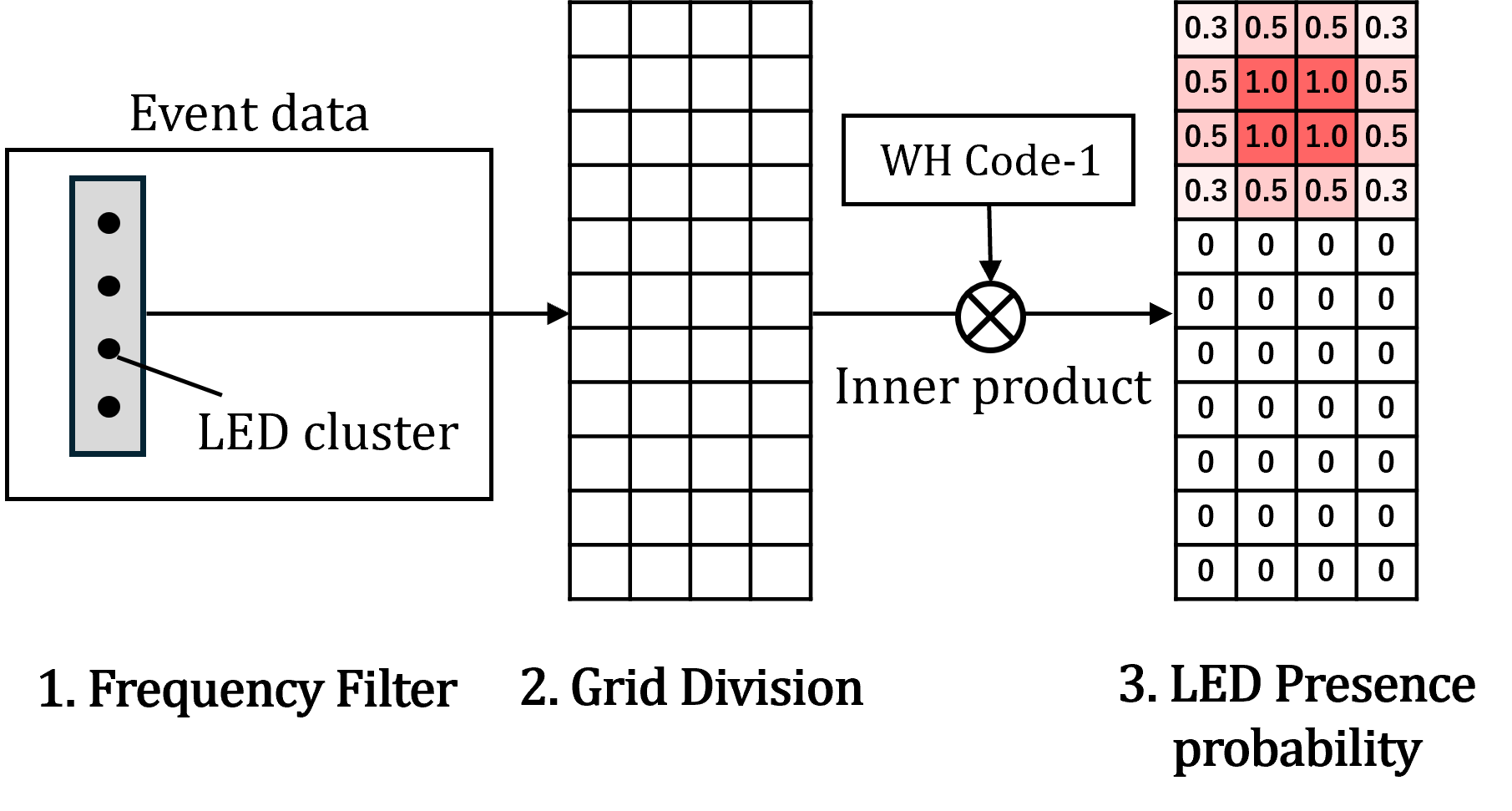}
    \vspace{-3mm}

\caption{Reciever Algorithm~\cite{sogasogasoga}}
\vspace{-3mm}
\label{fig:rec-algo}
\vspace{-1mm}
\end{figure}

\vspace{-1mm}
To enable simultaneous communication and ranging using multiple LED Clusters, it is essential to separate individual Clusters for both VLC and VLP functions. However, the LED bar used in this study consists of 96 LEDs densely arranged in a single vertical column measuring 96 cm in length, making it difficult to distinguish individual LEDs using conventional image processing techniques. Moreover, since the receiver is mounted on a moving vehicle, it must also track the LED bar as it shifts within the camera’s field of view during motion. To address these challenges, we propose an algorithm in which the transmitter continuously sends known pilot sequences, and the receiver calculates the cross-correlation between the received signal and the pilot sequence in each spatial region of the field of view to estimate and separate the LED presence probability distribution for each Cluster (Fig.~\ref{fig:rec-algo}).

\begin{figure}[ht]
    \centering
    \includegraphics[width = 0.80\linewidth]{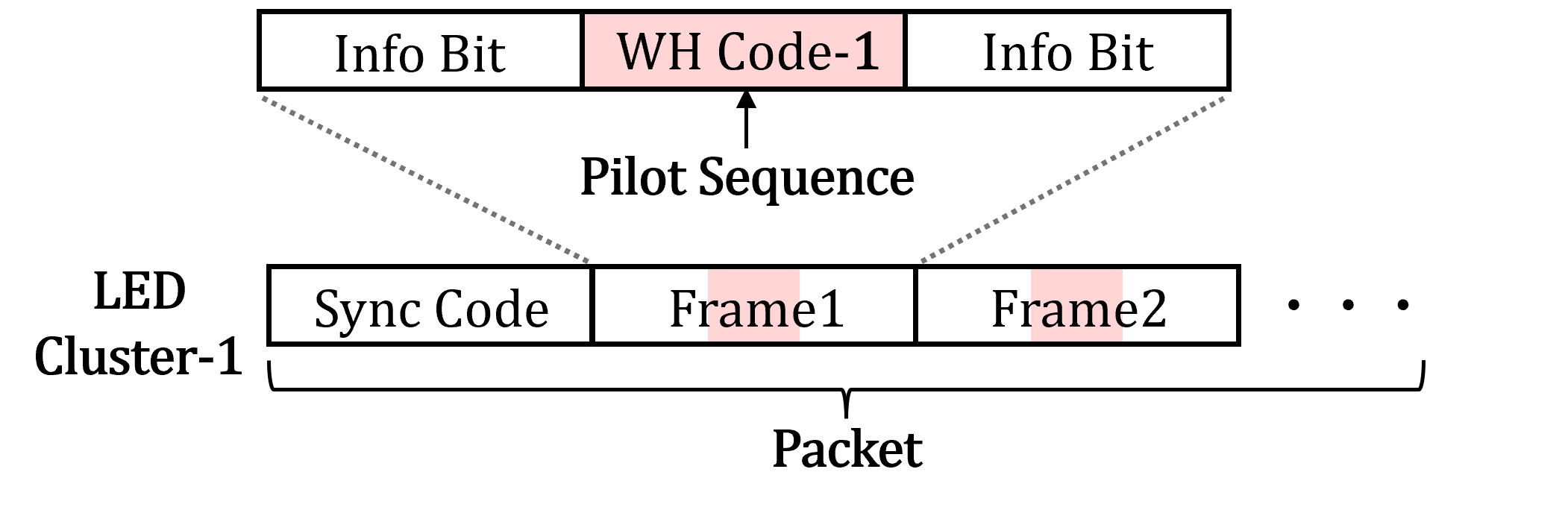}
    \vspace{-3mm}
    \caption{Packet Structure of Transmission Data~\cite{sogasogasoga}}
    \vspace{-4mm}
    \label{fig:Transmitter-packet}
\end{figure}

On the transmitter side, each LED Cluster is assigned a unique WH code ~\cite{Kiyasu1980} as its pilot sequence. These codes are transmitted using bipolar modulation with values in the range \([-1, 1]\), as illustrated in the transmission packet shown in Fig.~\ref{fig:Transmitter-packet}. WH codes exhibit mutual orthogonality, which suppresses interference from other Clusters during cross-correlation, thus enabling accurate identification. Furthermore, this study extends the number of assignable codewords by also using inverted WH codes, where each code is multiplied by -1. Although the inverted codes are not perfectly orthogonal to the original codes, their cross-correlation values are sufficiently low (typically below zero), which is acceptable for our algorithm that considers only regions with high correlation values as candidates for LED presence. This strategy enables a greater number of unique codes to be assigned even with shorter code lengths.

The use of WH spreading is motivated by the characteristics of event cameras. Since these cameras only generate events when luminance changes occur, transmitting a signal with continuous polarity (e.g., a long sequence of 1 or -1) would fail to produce detectable events, making signal reconstruction difficult. To avoid this, each information bit is spread using a bipolar WH code, introducing periodic polarity inversions in the modulated waveform. Let the spread bit sequence be defined as \( \mathbf{b} = [b_1, b_2, \ldots] \). By extracting every other bit where no polarity reversal occurs, we define a subset waveform \( \mathbf{b'} = [b'_1, b'_2, \ldots] \) (\ref{eq:1bitskip}). Moreover, in cases where bits are lost due to noise or event dropouts, the presence of inverted bits (i.e., bits multiplied by -1) can be exploited to recover the original value by reversing the polarity. The final reconstructed waveform, incorporating both subsets and inverted bit recovery, is denoted as \( \mathbf{B''} = [b''_1, b''_2, \ldots, b''_n] \) (\ref{eq:hokan}), offering robustness against noise and loss.
\vspace{-1mm}
\begin{equation}
 b'_i = b_{2i} (i = 1,2, \dots, n/2)
 \label{eq:1bitskip}
\end{equation}
\begin{equation}
b''_i =
\begin{cases}
b'_i & \text{if } b'_i \neq 0 \\
-b_{2i-1} & \text{if } b'_i = 0 \text{ and } i > 1
\end{cases}
\label{eq:hokan}
\vspace{-1mm}
\end{equation}

On the receiver side, the approximate position of the LED bar is first detected using a Frequency filter. The detected region is divided into grids composed of multiple pixels, and the events within each grid are aggregated for further processing. This grid-based approach enhances robustness to noise and event loss. Then, for each grid, the cross-correlation between the received data and the assigned WH code is computed, and the similarity to each LED Cluster is evaluated. Grids with high correlation values are interpreted as regions with a high probability of LED presence. Consequently, the resulting spatial distribution of correlation values can be directly interpreted as the probability distribution of LED presence.
In this algorithm, each pilot sequence is 16 bits long, and the information segment is also 16 bits, resulting in a total frame length of 32 bits. With a blinking frequency of 10,000 Hz, this configuration enables updates to the presence probability distribution approximately every 3.2 ms. While this method assumes a vertically aligned LED bar, it can be adapted to other layouts, such as square or two-column arrangements, by modifying the grid division and correlation process accordingly.

\vspace{-2mm}
\subsection{VLC Decoding}
\label{method:decoding}
Based on the LED presence probability distribution described in Section A, the demodulation process for VLC is performed. Specifically, for a given LED Cluster \( k \), the presence probability in each grid within the field of view is denoted as \( w^{(k)}_{x,y} \). Only the grids whose probability exceeds a predefined threshold are considered. For each of these grids, the information bit segment \(\mathbf{i}^{(k)}_{x,y}(t) \) is weighted by \( w^{(k)}_{x,y} \), and all weighted values are summed to obtain an integrated information bit \( \mathbf{I}^{(k)}(t) \) at time \( t \) (Eq.~\ref{eq:ave}).

\vspace{-2mm}
\begin{equation}
    \mathbf{I}^{(k)}(t) = \sum_{x,y} w_{x,y}^{(k)} \mathbf{i}_{x,y}(t)
\label{eq:ave}
\vspace{-2mm}
\end{equation}

The resulting \( I^{(k)}(t) \) is then decoded using a method based on the WH transform. This decoding technique utilizes the orthogonality property of WH codes to calculate the similarity between the received bit sequence and each possible codeword, selecting the one with the highest similarity as the decoded signal~\cite{Ehara}. This method is suited to the characteristics of output data and also provides error-correction capabilities.

This approach remains effective even when the receiver is mounted on a moving vehicle and subject to vibration. The timing deviation of the information bit with respect to the transmission start of the pilot sequence is no more than approximately 0.8 ms, which limits the impact of short-term motion. Previous research~\cite{kinoshita} reported that, when using a lens with a focal length of 35 mm, the maximum displacement observed is approximately 1.5 pixels per millisecond during driving at a speed of 30 km/h. Since our method processes signals on a per-grid basis, where each grid spans at least two pixels, the algorithm remains robust against such small displacements.

However, when the distance between the transmitter and receiver becomes large, the number of pixels representing the transmitter in the field of view decreases. Consequently, each grid contains less information, and the effect of motion-induced errors becomes more pronounced. This limitation has been observed in our experiments and remains a challenge to be addressed in future improvements.
\vspace{-3mm}
\subsection{Measurement and Phase Only Correlation}
\vspace{-2mm}
\label{method:measurement}
Finally, based on the LED Cluster presence probability distribution calculated in Section A, we describe the ranging algorithm used for VLP (Visible Light Positioning)~\cite{ohmura, kobayashi}. This method estimates the distance between the transmitter and receiver with high accuracy by performing triangulation based on the spatial separation of LED Clusters and the camera's intrinsic parameters.

First, for each LED Cluster, a pixel-level probability distribution is generated. Specifically, we apply a predefined threshold to the grid-level presence probabilities \( w^{(k)}_{x,y} \) and extract only the grids with high probability values. The pixels within the selected grids are then re-evaluated to compute the pixel-level presence probabilities \( w'^{(k)}_{x',y'} \), which are then arranged into a two-dimensional array. This process suppresses background noise while preserving a high-resolution image representation suitable for POC.

Next, we select two probability distribution maps corresponding to a pair of LED Clusters and apply POC and triangulation to estimate the distance. By applying POC to these two distributions, the relative distance \( l \) (in pixels) between the LED Clusters in the field of view is calculated. For further accuracy, we refine the distance by fitting a sinc function to the correlation peak and estimating the sub-pixel peak location, resulting in a more precise inter-cluster distance \( l' \). This procedure is repeated across multiple pairs of LED Clusters to obtain multiple distance estimates.

Using the sub-pixel estimated distance \( l' \), along with the pixel pitch \( \alpha \), focal length \( f \), and known physical distance \( S \) between the two LED Clusters, we compute the actual distance \( L \) between the transmitter and receiver via triangulation (Eq.~\ref{eq:sankaku}). This triangulation-based approach is particularly effective for achieving accurate ranging even at long distances.
\vspace{-1mm}
\begin{equation}
    L = \frac{fS}{l'\alpha}
\label{eq:sankaku}
\vspace{-1mm}
\end{equation}

Finally, the estimated distances \( L \) from all LED Cluster pairs are aggregated to produce a final ranging result. To mitigate the effect of outliers, the maximum and minimum values are excluded, and an interquartile range (IQR) method is applied to remove statistical anomalies. Additionally, since shorter distances between LED Clusters are more susceptible to measurement error, we assign weights based on the inter-cluster spacing and compute a weighted average. With event cameras, intended events may fail to trigger, while spurious events can be generated; this occurs especially frequently for negative events on the IMX636 sensor. These anomalies influence the estimated distance. Applying outlier removal in conjunction with weighted averaging effectively mitigates the problem, thereby improving both the stability and accuracy of the ranging results.

\vspace{-2mm}

\section{Field Experiments}
\label{sec:experiment}
\vspace{-2mm}
To evaluate the system proposed in this study, experiments were conducted in an outdoor vehicle-mounted environment. Verification was conducted in a driving environment by mounting an event camera with a 35 mm lens on the vehicle and placing the LED bar 1.5 meters to the left of the vehicle. The experiment was performed under dynamic conditions, simulating small-mobility scenarios with vehicle speeds of 20 km/h (5.6 m/s), 30 km/h (8.3 m/s), and 40 km/h (11.1 m/s). The LED Clusters were configured by grouping every six LEDs, resulting in 16 LED Clusters in total. Under this configuration, the communication rate was 27 kbps.

\begin{figure}[ht]
    \centering
    \includegraphics[width = 0.80\linewidth]{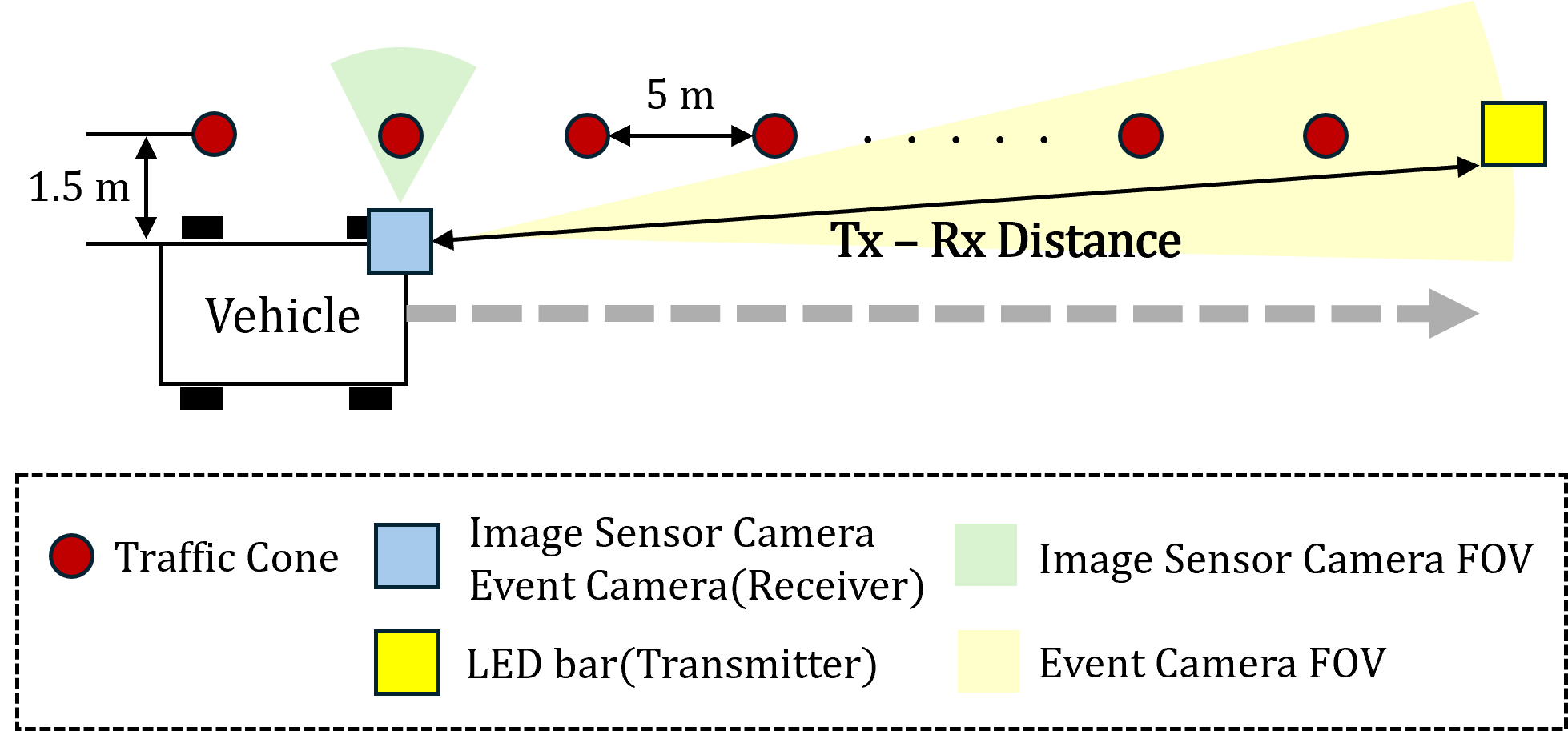}
    \caption{Experimental setup }
    \vspace{-4mm}
    \label{fig:exp}
\end{figure}

For distance measurement, the outermost Cluster was excluded due to instability caused by defocusing, and the central Cluster was also omitted. Instead, pairs of Clusters with longer inter-cluster spacing were selected for triangulation. To obtain the ground truth for distance measurement, we prepared an experimental setup as shown in Fig.~\ref{fig:exp}. First, pylons were placed at 5-meter intervals along a line located 1.5 meters to the side of the receiver. These pylons were recorded from inside the vehicle using an image sensor camera synchronized with the event camera. By analyzing the video captured by the image sensor, we estimate the time-varying distance between the transmitter and receiver.

\begin{figure}[ht]
    \centering
    \includegraphics[width = 0.9\linewidth]{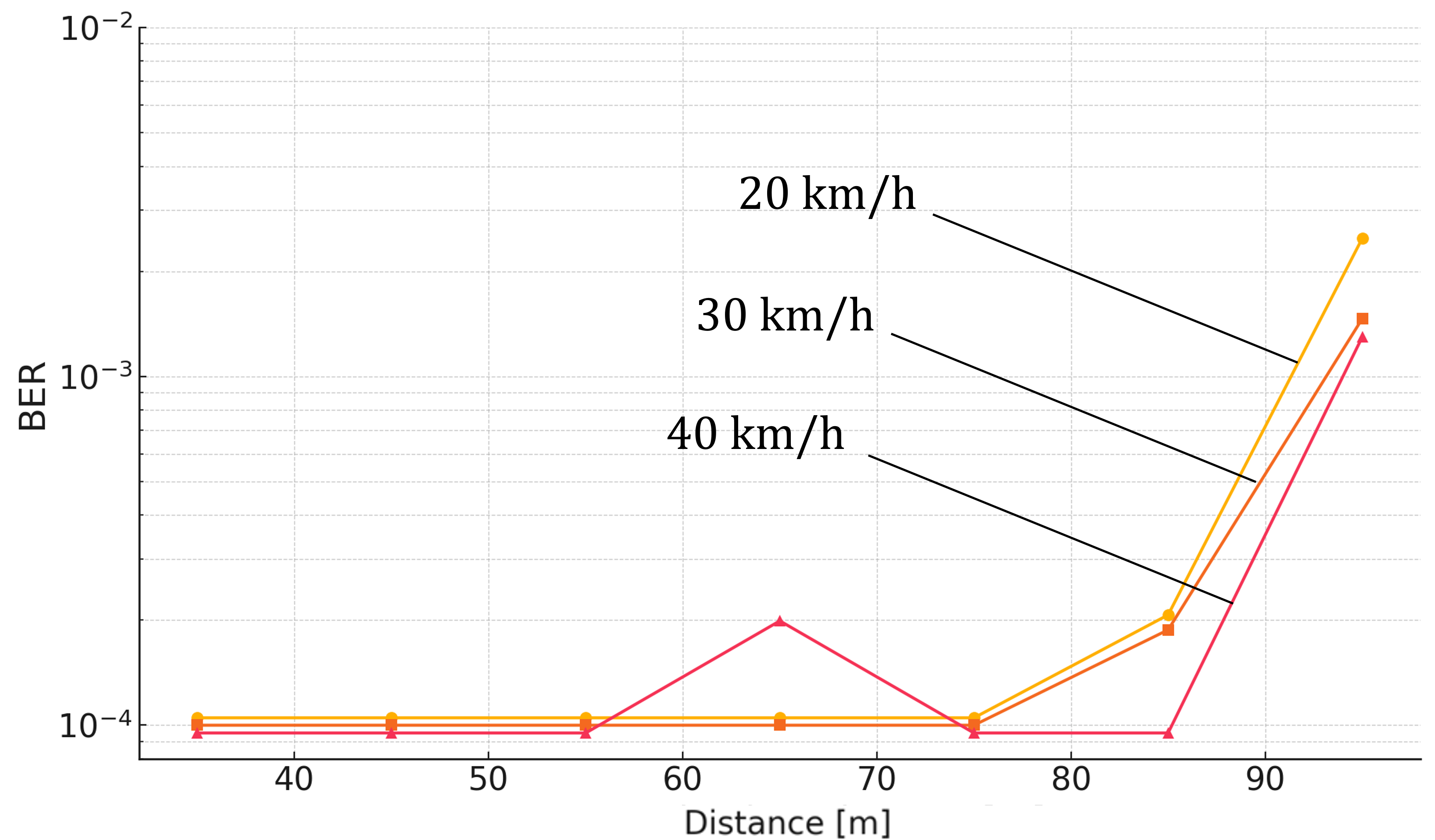}
    \caption{BER-Distance Characteristics in a Vehicular Environment}
    \vspace{-5mm}
    \label{fig:VLCresult}
\end{figure}

\begin{figure}[ht]
    \centering
    \includegraphics[width = 0.9\linewidth]{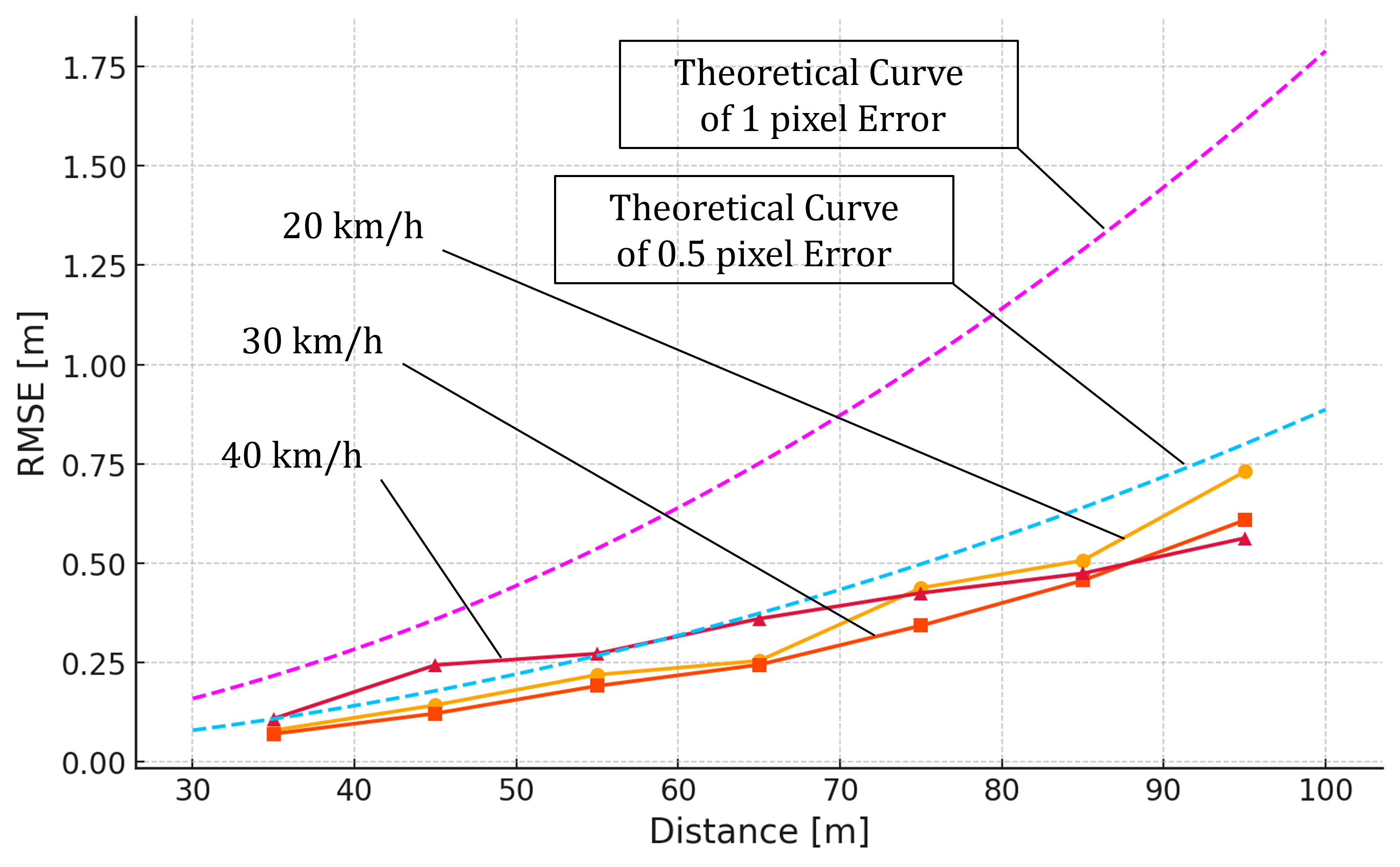}
    \caption{RMSE Characteristics in a Vehicular Environment}
    \vspace{-5mm}
    \label{fig:VLPresult}
\end{figure}

The experimental results are presented in Figs.~\ref{fig:VLCresult}, \ref{fig:VLPresult}. First, Fig.~\ref{fig:VLCresult} shows the results of VLC. For clarity, the bit error rate (BER) was averaged over 10-meter segments and plotted at the midpoint of each segment. The results demonstrate that the BER remains below 0.01 across all speed levels for distances ranging from 30 m to 100 m. This indicates that error-free communication can be achieved by incorporating error-correcting codes such as Polar codes.

Next, Fig.~\ref{fig:VLPresult} presents the results of VLP. Distance estimation was performed each time a pilot sequence was received, and RMSE between the estimated and ground-truth values was computed over each 10-meter segment and plotted at the segment midpoint. Additionally, the theoretical errors introduced by POC-based distance estimation, which occur when there is a 1-pixel or 0.5-pixel deviation in the image plane, were also plotted as reference curves. From these results, it is evident that distance estimation was achieved with sub-pixel accuracy over the entire 30 - 100 m range. At a distance of around 30 meters, a ranging accuracy of approximately less than 10 cm is achieved. At that time, the scan rate was 290 Hz.

Table ~\ref{tab:rmse_comparison} presents a comparison between the proposed method and a conventional method that applied POC for distance estimation using two LED sources ~\cite{kobayashi}. In a previous study, the experiment was conducted using two LED clusters, each consisting of five LEDs arranged in a single spatial cluster and positioned so that they did not interfere with other clusters. The ranging was performed by applying POC to 3 ms worth of event data collected during motion at a speed of 30 km/h. RMSE values in the 30–40 m and 40–50 m distance intervals were used as the primary metrics for evaluation. As shown in Table ~\ref{tab:rmse_comparison}, the proposed method achieves a significant improvement in accuracy, reducing the RMSE by approximately 70\% compared to the prior method. This improvement can be attributed to the introduction of a weighted averaging process across multiple ranging results, which contributes to increased robustness and accuracy. 
\vspace{-2mm}
\begin{table}[htbp]
  \centering
  \caption{Comparison of RMSE [m] between Proposed Method and 2LED Cluster at 30\,km/h}
  \label{tab:rmse_comparison}
  {\small
  \begin{tabularx}{\linewidth}{lXX}
    \toprule
    \textbf{Distance Range} & \textbf{Proposed Method [m]} & \textbf{Conventional Method [m]} \\
    \midrule
    30--40\,m & 0.0702 & 0.2819 \\
    40--50\,m & 0.1216 & 0.4182 \\
    \bottomrule
  \end{tabularx}
  }
  \vspace{-3mm}
\end{table}

We compare the high-speed-camera VLP/VLC system reported by Ohmura \cite{ohmura} with the event-camera-based system proposed in this study. Because Yamazato et al. do not provide a communication-rate figure, we rely on Ruirui’s survey \cite{ruirui}, which cites a high-speed-camera scheme achieving 32 kbps with a 32 × 32 LED array. To ensure a fair comparison, we normalize the results by the number of transmitter LEDs and evaluate the data rate per LED; the outcomes are summarized in Table \ref{tab:comparison}. The event-camera system exhibits a higher per-LED throughput, giving it an advantage when transmitters of identical physical size are assumed. This benefit can be attributed to the event camera’s superior temporal resolution.

Next, we compare ranging accuracy. In the 30–60 m range, the proportion of measurements with a distance error of 0.5 m or more was 8.5 \% for the high-speed-camera system and 0.4 \% for the event-camera system. This improvement arises from applying outlier rejection and averaging techniques that suppress the inherent variability in event generation.

In summary, the proposed event-camera-based system surpasses conventional high-speed-camera approaches in both VLC throughput and VLP ranging performance.

\vspace{-2mm}
\begin{table}[htbp]
  \centering
  \caption{Comparison of Communication and Ranging Performance between High-Speed Cameras and Event Cameras}
  \label{tab:comparison}
  {\small
  \begin{tabularx}{\linewidth}{lXX}
    \toprule
    \textbf{} & \textbf{Event Camera} & \textbf{High-Speed Camera} \\
    \midrule
    Per-LED Data Rate & 281 bps & 31 bps \\
    Ranging Error & 0.4\% & 8.5\% \\
    \bottomrule
  \end{tabularx}
  }
  \vspace{-3mm}
\end{table}

\vspace{-2mm}
\section{Conclusion}
\vspace{-2mm}

\label{sec:conclution}
In this study, we proposed, for the first time, a vehicle-mounted system that simultaneously performs ranging and communication using an event camera. Unlike conventional approaches that separately transmit LED blinking patterns for VLC and VLP, our system enables simultaneous execution of both functions by sharing the LED Cluster presence probability distribution derived from pilot sequences. Furthermore, by utilizing three or more LED Clusters, we aggregated multiple pairwise distance estimates using a weighted averaging method, achieving high accuracy and robustness.

We conducted field experiments by mounting the receiver on a vehicle and evaluating system performance over distances ranging from 30 to 100 meters. The results demonstrated that the VLC function achieved a BER below 0.01, enabling error-free communication when combined with error correction codes. For VLP, the system achieved high-accuracy ranging with estimation errors of less than one pixel in most cases. Additionally, a high scan rate of approximately 300 Hz was successfully achieved for real-time ranging.

On the other hand, the current system is limited to estimating "distance" rather than the full "position" of the vehicle in three-dimensional space. To overcome this limitation, we consider introducing multi-transmitter configurations and multi-point localization techniques as a promising future direction.

The experiments so far were limited to 40 km/h; beyond this speed, vibration degrades tracking, so more precise algorithms are required. In addition, event cameras’ high dynamic range lets them handle overexposed scenes, so the system should also be tested in High Dynamic Range and strong-ambient-light conditions.


\end{document}